\documentclass{article}

\usepackage[final]{nips_2016}

\usepackage[utf8]{inputenc} 
\usepackage[T1]{fontenc}    
\usepackage{url}            
\usepackage{booktabs}       
\usepackage{amsfonts}       
\usepackage{nicefrac}       
\usepackage[
        activate={true,nocompatibility},
        tracking=true
        kerning=true,
        spacing=true
]{microtype}      
\usepackage{graphicx}
\usepackage{algorithm}
\usepackage{algorithmic}

\usepackage{amssymb}

\newcommand{\disc}{\mathrm{Disc}}
\newcommand{\enc}{\mathrm{Enc}}
\newcommand{\dec}{\mathrm{Dec}}
\newcommand{\gen}{\mathrm{Gen}}
\newcommand{\id}{\mathrm{id}}
\newcommand{\adv}{\mathrm{Adv}}

\newcommand{\given}{\;|\;}
\newcommand{\KL}{\mathrm{KL}}

\title{Disentangling factors of variation in deep representations using adversarial training}

\author{
	Michael Mathieu, Junbo Zhao, Pablo Sprechmann, Aditya Ramesh, Yann LeCun \\
	719 Broadway, 12th Floor, New York, NY 10003 \\
	\texttt{\{mathieu, junbo.zhao, pablo, ar2922, yann\}@cs.nyu.edu}  \\
}

\begin{document}

\maketitle

\begin{abstract}

We introduce a conditional generative model for learning to disentangle the
hidden factors of variation within a set of labeled observations, and separate
them into complementary codes. One code summarizes the \emph{specified} factors
of variation associated with the labels. The other summarizes the remaining
\emph{unspecified} variability. During training, the only available source of
supervision comes from our ability to distinguish among different observations
belonging to the same class. Examples of such observations include images of a
set of labeled objects captured at different viewpoints, or recordings of set
of speakers dictating multiple phrases. In both instances, the intra-class
diversity is the source of the unspecified factors of variation: each object is
observed at multiple viewpoints, and each speaker dictates multiple phrases.
Learning to disentangle the specified factors from the unspecified ones becomes
easier when strong supervision is possible. Suppose that during training, we
have access to pairs of images, where each pair shows two different objects
captured from the same viewpoint. This source of alignment allows us to solve
our task using existing methods. However, labels for the unspecified
factors are usually unavailable in realistic scenarios where data acquisition
is not strictly controlled. We address the problem of disentaglement in this
more general setting by combining deep convolutional autoencoders with a form of
adversarial training. Both factors of variation are implicitly captured in the
organization of the learned embedding space, and can be used for solving
single-image analogies. Experimental results on synthetic and real datasets
show that the proposed method is capable of generalizing to unseen classes and
intra-class variabilities.

\end{abstract}

\section{Introduction}

A fundamental challenge in understanding sensory data is learning to
disentangle the underlying factors of variation that give rise to the
observations~\cite{bengio2009learning}. For instance, the factors of variation
involved in generating a speech recording include the speaker's attributes,
such as gender, age, or accent, as well as the intonation and words being
spoken. Similarly, the factors of variation underlying the image of an object
include the object's physical representation and the viewing conditions. The
difficulty of disentangling these hidden factors is that, in most real-world
situations, each can influence the observation in a different and unpredictable
way. It is seldom the case that one has access to rich forms of labeled data in
which the nature of these influences is given explicitly.

Often times, the purpose for which a dataset is collected is to further
progress in solving a certain supervised learning task. This type of learning
is driven completely by the labels. The goal is for the learned representation
to be invariant to factors of variation that are uninformative to the task at
hand. While recent approaches for supervised learning have enjoyed tremendous
success, their performance comes at the cost of discarding sources of variation
that may be important for solving other, closely-related tasks. Ideally, we would
like to be able to learn representations in which the uninformative factors of
variation are separated from the informative ones, instead of being discarded.

Many other exciting applications require the use of generative models that are
capable of synthesizing novel instances where certain key factors of variation
are held fixed. Unlike classification, generative modeling requires preserving all factors of
variation. But merely preserving these factors is not sufficient for many tasks
of interest, making the disentanglement process necessary. For example, in
speech synthesis, one may wish to transfer one person's dialog to another
person's voice. Inverse problems in image processing, such as denoising and
super-resolution, require generating images that are perceptually consistent
with corrupted or incomplete observations. 

In this work, we introduce a deep conditional generative model that learns to
separate the factors of variation associated with the labels from the other
sources of variability. We only make the weak assumption that we are able to
distinguish between observations assigned to the same label during training. 
To make disentanglement possible in this more general setting, we leverage both
Variational Auto-Encoders~(VAEs)~\cite{kingma2013auto,rezende2014stochastic}
and Generative Adversarial Networks~(GANs)~\cite{Goodfellow2014adversarial}.

\section{Related work}
\label{sec.related.work}

There is a vast literature on learning disentangled representations. Bilinear
models~\cite{tenenbaum2000} were an early approach to separate content and
style for images of faces and text in various fonts. What-where
autoencoders~\cite{ranzato2007,swwae2016} combine discrimination and
reconstruction criteria to attempt to recover the factors of variation not
associated with the labels. In~\cite{hinton2011capsules}, an autoencoder is
trained to separate a translation invariant representation from a code that is
used to recover the translation information. 
In~\cite{cheung2014}, the authors show that
standard deep architectures can discover and explicitly represent factors of
variation aside those relevant for classification, by combining autoencoders
with simple regularization terms during the training. 
In the context of generative models, the work in~\cite{reed2014learning}
extends the Restricted Boltzmann Machine by partitioning its hidden state into
distinct factors of variation. The work presented in~\cite{kingma2014semi} uses
a VAE in a semi-supervised learning setting. Their approach is able to
disentangle the label information from the hidden code by providing an
additional one-hot vector as input to the generative model. Similarly,
\cite{advautoencoder} shows that autoencoders trained in a semi-supervised
manner can transfer handwritten digit styles using a decoder conditioned on a
categorical variable indicating the desired digit class. The main difference
between these approaches and ours is that the former cannot generalize to
unseen identities.

The work in~\cite{chairs2014,kulkarni2015deep} further explores the application
of content and style disentanglement to computer graphics. Whereas computer
graphics involves going from an abstract description of a scene to a rendering,
these methods learn to go backward from the rendering to recover the abstract
description. This description can include attributes such as orientation and
lighting information. While these methods are capable of producing impressive
results, they benefit from being able to use synthetic data, making strong
supervision possible.

Closely related to the problem of disentangling factors of variations in
representation learning is that of learning fair
representations~\cite{fairae,edwards2015censoring}. In particular, the Fair
Variational Auto-Encoder~\cite{fairae} aims to learn representations that are
invariant to certain nuisance factors of variation, while retaining as much of
the remaining information as possible. The authors propose a variant of the VAE
that encourages independence between the different latent factors of
variation. 

The problem of disentangling factors of variation also plays an important role
in completing image analogies, the goal of the end-to-end model proposed
in~\cite{spritespaper}. Their method relies on having access to matching
examples during training. 
%
%
Our approach requires neither matching observations nor labels aside from the
class identities. These properties allow the model to be trained on data with a
large number of labels, enabling generalizing over the classes present in the
training data.

\section{Background}
\label{sec.background}

\subsection{Variational autoencoder}
\label{sec.vae}

The VAE framework is an approach for modeling a data distribution using a
collection of independent latent variables. Let $x$ be a random variable (real
or binary) representing the observed data and $z$ a collection of real-valued
latent variables. The generative model over the pair $(x, z)$ is given by $p(x,
z) = p(x \given z) p(z)$, where $p(z)$ is the prior distribution over the
latent variables and $p(x \given z)$ is the conditional likelihood function.
Generally, we assume that the components of $z$ are independent Bernoulli or
Gaussian random variables. The likelihood function is parameterized by a deep
neural network referred to as the \emph{decoder.} 



A key aspect of VAEs is the use of a learned approximate inference procedure
that is trained purely using gradient-based methods
\cite{kingma2013auto,rezende2014stochastic}. This is achieved by using a
learned approximate posterior $q(z \given x)= N(\mu, \sigma I)$ whose
parameters are given by another deep neural network referred to as the
\emph{encoder.} Thus, we have $z \sim \enc(x) = q(z|x)$ and $\tilde{x} ~
\dec(z) = p(x|z)$. The parameters of these networks are optimized by minimizing
the upper-bound on the expected negative log-likelihood of $x$, which is given by
\begin{equation}
	\mathbb{E}_{q(z \given x)}[-\log p_{\theta}(x \given z)]  + \KL(q(z|x) \;||\; p(z)).
	\label{eq.vae}
\end{equation}
The first term in~(\ref{eq.vae}) corresponds to the reconstruction error, and
the second term is a regularizer that ensures that the approximate posterior
stays close to the prior. 

\subsection{Generative adversarial networks}
\label{sec.gan}

Generative Adversarial Networks~(GAN)~\cite{Goodfellow2014adversarial} have
enjoyed great success at producing realistic natural images~\cite{DCGAN}. The
main idea is to use an auxiliary network $\disc$, called the
\emph{discriminator}, in conjunction with the generative model, $\gen$. The
training procedure establishes a min-max game between the two networks as
follows. On one hand, the discriminator is trained to differentiate between
natural samples sampled from the true data distribution, and synthetic images
produced by the generative model. On the other hand, the generator is trained
to produce samples that confuse the discriminator into mistaking them for
genuine images. The goal is for the generator to produce increasingly more
realistic images as the discriminator learns to pick up on increasingly more
subtle inaccuracies that allow it to tell apart real and fake images.

Both $\disc$ and $\gen$ can be conditioned on the label of the input that we
wish to classify or generate, respectively~\cite{mirza2014conditional}. This
approach has been successfully used to produce samples that belong to a
specific class or possess some desirable
property~\cite{denton2015deep,mathieu2016,DCGAN}. The training objective can be
expressed as a min-max problem given by
\begin{equation}
	\min_\gen \max_\disc L_{\mathrm{gan}}, \quad\textrm{where}\quad 
	L_{\textrm{gan}} =  \log\disc(x, \id) + \log(1 - \disc(\gen(z, \id), \id)).
\label{eq.gan}
\end{equation}
where $p_d(x, id)$ is the data distribution conditioned on a given class label
$id$, and $p(z)$ is a generic prior over the latent space (e.g. $N(0, I)$).



\section{Model}
\label{src.model}


\subsection{Conditional generative model}
\label{sec.cond_model}

We introduce a conditional probabilistic model admitting two independent
sources of variation: an observed variable $s$ that characterizes the specified
factors of variation, and a continuous latent variable $z$ that characterizes
the remaining variability. The variable $s$ is given by a vector of real
numbers, rather than a class ordinal or a one-hot vector, as we intend for the
model to generalize to unseen identities. 

Given an observed specified component $s$, we can sample
\begin{equation}
	z \sim p(z) = N(0, I) \quad\textrm{and}\quad x \sim p_{\theta}(x \given z, s),
	\label{eq.model}
\end{equation}
in order to generate a new instance $x$ compatible with $s$.

The variables $s$ and $z$ are marginally independent, which promotes
disentanglement between the specified and unspecified factors of variation.
Again here, $p_{\theta}(x|z,s)$ is a likelihood function described by and
decoder network, $\dec$, and the approximate posterior is modeled using an
independent Gaussian distribution, $q_{\phi}(z|x,s)=N(\mu,\sigma I)$, whose
parameters are specified via an encoder network, $\enc$. In this new setting,
the variational upper-bound is be given by
\begin{equation}
	\mathbb{E}_{q(z \given x, s)}[-\log p_{\theta}(x \given z, s)] +
		\KL(q(z \given x, s) \given p(z)).
	\label{eq.new_vae}
\end{equation}

The specified component $s$ can be obtained from one or more images belonging
to the same class. In this work, we consider the simplest case in which $s$ is
obtained from a single image. To this end, we define a deterministic encoder
$f_s$ that maps images to their corresponding specified components. All sources
of stochasticity in $s$ come from the data distribution. The conditional
likelihood given by~(\ref{eq.model}) can now be written as $x \sim p_{\theta}(x
\given z,f_{s}(x'))$ where $x'$ is any image sharing the same label as $x$,
including $x$ itself. In addition to $f_s$, the model has an additional encoder
$f_z$ that parameterizes the approximate posterior $q(z \given x, s)$. It is
natural to consider an architecture in which parameters of both encoders are
shared.


We now define a single encoder $\enc$ by $\enc(x) = (f_s(x), f_z(x)) = (s,
(\mu, \sigma) = (s, z)$, where $s$ is the specified component, and $z = (\mu,
\sigma)$ the parameters of the approximate posterior that constitute the
unspecified component. To generate a new instance, we synthesize $s$ and $z$
using $\dec$ to obtain $\tilde{x} = \dec(s, z)$.


The model described above cannot be trained by minimizing the log-likelihood
alone. In particular, there is nothing that prevents all of the information
about the observation from flowing through the unspecified component. The
decoder could learn to ignore $s$, and the approximate posterior could map
images belonging to the same class to different regions of the latent space.
This degenerate solution can be easily prevented when we have access to labels
for the unspecified factors of variation, as in~\cite{spritespaper}. In this
case, we could enforce that $s$ be informative by requiring that $\dec$ be able
to reconstruct two observations having the same unspecified label after their
unspecified components are swapped. But for many real-world scenarios, it is
either impractical or impossible to obtain labels for the unspecified factors
of variation. In the following section, we explain a way of eliminating the
need for such labels.


\subsection{Discriminative regularization}

An alternative approach to preventing the degenerate solution described in the
previous section, without the need for labels for the unspecified components,
makes use of GANs~(\ref{sec.gan}). As before, we employ a procedure in which
the unspecified components of a pair of observations are swapped. But since the
observations need not be aligned along the unspecified factors of variation, it
no longer makes sense to enforce reconstruction. After swapping, the class
identities of both observations will remain the same, but the sources of
variability within their corresponding classes will change. Hence, rather than
enforcing reconstruction, we ensure that both observations are assigned high
probabilities of belonging to their original classes by an external
discriminator. Formally, we introduce the discriminative term given
by~(\ref{eq.gan}) into the loss given by~(\ref{eq.new_vae}), yielding
\begin{equation}
	\mathbb{E}_{q(z \given x, s)}[-\log p_{\theta}(x \given z, s)] +
		\KL(q(z \given x, s) \;||\; p(z)) + \lambda L_{\textrm{gan}},
\label{eq.new_vae}
\end{equation}
where $\lambda$ is a non-negative weight.


Recent works have explored combining VAE with GAN \cite{larsen2015autoencoding,dumoulin2016adversarially}. 
%
These approaches aim at including a recognition network (allowing solving inference problems) to the GAN framework. 
%
In the setting used in this work, GAN is used to compensate the lack of aligned training data. 
The work in \cite{larsen2015autoencoding} investigates the use of GANs for obtaining perceptually better loss functions (beyond pixels). While this is not the goal of our work, our framework is able to generate sharper images, which comes as a side effect. We evaluated including a GAN loss also for samples, however, the system became unstable without leading to perceptually better generations. An interesting variant could be to use separate discriminator for images generated with and without supervision.

\subsection{Training procedure}
\label{sec.training}

Let $x_1$ and $x'_1$ be samples sharing the same label, namely $id_1$, and $x_2$ a sample belonging to a different class, $id_2$. 
On one hand we want to minimize the upper bound of negative log likelihood of $x_1$ when feeding to the decoder inputs of the form $(z_1, f_s(x_1))$ and $( z_1, f_s(x'_1))$, where $z_1$ are samples form the approximate posterior $q(z|x_1)$. 
On the other hand, we want to minimize the adversarial loss of samples generated by feeding to the decoder inputs given by $(z, f_s(x_2))$, where $z$ is sampled from the approximate posterior $q(z|x_1)$. This corresponds to swapping specified and unspecified factors of $x_1$ and $x_2$. We could only use upper bound if we had access to aligned data.
As in the GAN setting described in Section~\ref{sec.gan}, we alternate this procedure with updates of the adversary network. The diagram of the network is shown in figure~\ref{figure:network_diagram}, and the described training procedure is summarized in on Algorithm 1, in the supplementary material.

\begin{figure}[ht]
  \centering
  \includegraphics[width=0.7\linewidth]{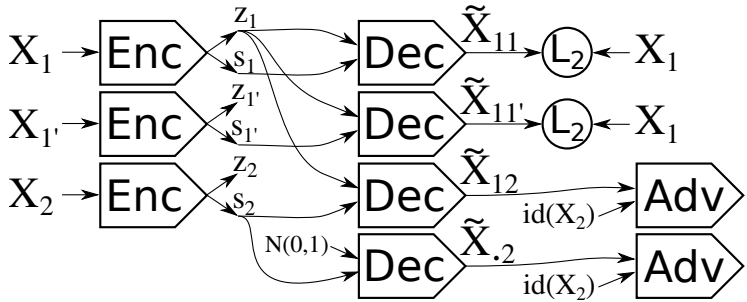}
  \caption{\label{figure:network_diagram}Training architecture. The inputs $x_1$ and $x_1'$ are two different samples with the same label, whereas $x_2$ can have any label.}
\vspace{-3ex}
\end{figure}

\section{Experiments}
\label{sec.exp}

{\bf Datasets.} We evaluate our model on both synthetic and real datasets: Sprites dataset~\cite{spritespaper}, MNIST~\cite{lecun1998gradient}, NORB~\cite{lecun2004norb} and the Extended-YaleB dataset~\cite{georghiades2001few}. We used Torch7 ~\cite{collobert2011torch7} to conduct all experiments. The network architectures follow that of DCGAN~\cite{DCGAN} and are described in detail in the supplementary material.

{\bf Evaluation.} %
To the best of our knowledge, there is no standard benchmark dataset (or task) for evaluating disentangling performance \cite{cheung2014}.
%
We propose two forms of evaluation to illustrate the behavior of the proposed framework, one qualitative and one quantitative.

Qualitative evaluation is obtained by visually examining the perceptual quality of single-image analogies and conditional images generation. 
For all datasets, we evaluated the models in four different settings: 
{\it swapping:} given a pair of images, we generate samples conditioning on the specified component extracted from one of the images and sampling from the approximate posterior obtained from the other one. This procedure is analogous to the sampling technique employed during training, described in Section~\ref{sec.training}, and corresponds to solving single-image analogies; 
{\it retrieval:} in order to asses the correlation between the specified and unspecified components, we performed nearest neighbor retrieval in the learned embedding spaces. We computed the corresponding representations for all  samples (for the unspecified component we used the mean of the approximate posterior distribution) and then retrieved the nearest neighbors for a given query image; 
{\it interpolation:} to evaluate the coverage of the data manifold, we generated a sequence of images by linearly interpolating the codes of two given test images (for both specified and unspecified representations); 
{\it conditional generation:} given a test image, we generate samples conditioning on its specified component, sampling directly from the prior distribution, $p(z)$.
In all the experiments images were randomly chosen from the test set, please see specific details for each dataset. 

The objective evaluation of generative models is a difficult task and itself subject of current research \cite{theis2015note}. 
Frequent evaluation metrics, such as measuring the log-likelihood of a set of validation samples, are often not very meaningful as they do not correlate to the perceptual quality of the images \cite{theis2015note}. Furthermore, the loss function used by our model does not correspond a bound on the likelihood of a generative model, which would render this evaluation less meaningful. 
As a quantitative measure, we evaluate the degree of disentanglement via a classification task. Namely, we measure how much information about the identity is contained in the specified and unspecified components.

\begin{figure}[t]
\centering
\minipage{0.40\textwidth}
\centering
\includegraphics[width=0.8\columnwidth]{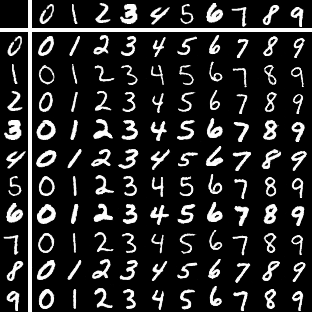}
\endminipage 
\minipage{0.40\textwidth}
\centering
\includegraphics[width=0.8\columnwidth]{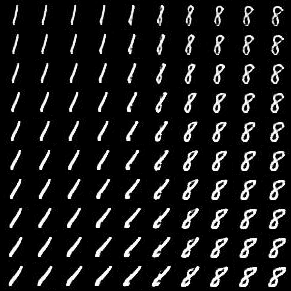}
\endminipage
\caption{left(a): A visualization grid of 2D MNIST image swapping generation. The top row and leftmost column digits come from the test set. The other digits are generated using $z$ from leftmost digit, and $s$ from the digit at the top of the column. The diagonal digits show reconstructions. 
Right(b): Interpolation visualization. Digits located at top-left corner and bottom-right corner come from the dataset. The rest digits are generated by interpolating $s$ and $z$. Like (a), each row has constant a $z$ each column a constant $s$.}
\label{fig:mnist}
\vspace{-1ex}
\end{figure}

{\bf MNIST.} 
In this setup, the specified part is simply the class of the digit. 
The goal is to show that the model is able to learn to disentangle the style from the identity of the digit and to produce satisfactory analogies.
We cannot test the ability of the model to generalize to unseen identities. In this case, one could directly condition on a class label \cite{kingma2014semi,advautoencoder}. It is still interesting that the proposed model is able to transfer handwriting style without having access to matched examples while still be able to learn a smooth representation of the digits as show in the interpolation results.
Results are shown in Figure \ref{fig:mnist}. We can see that both swapping and interpolation give very good results. 

\begin{figure}[t]
\centering
\minipage{0.40\textwidth}
\centering
	\includegraphics[width=0.8\columnwidth]{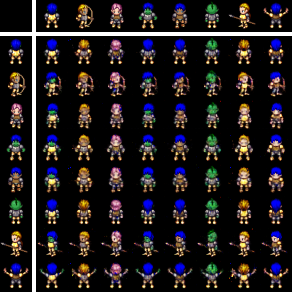}
\endminipage 
\minipage{0.40\textwidth}
\centering
\includegraphics[width=0.8\columnwidth]{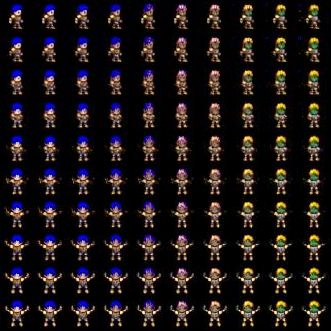}
\endminipage
\caption{left(a): A visualization grid of 2D sprites swapping generation. Same visualization arrangement as in \ref{fig:mnist}(a); right(b): Interpolation visualization. Same arrangement as in \ref{fig:mnist}(b).}
\label{fig:sprite}
\vspace{-1ex}
\end{figure}

\begin{figure}[t]
\centering
\minipage{0.40\textwidth}
\centering
\includegraphics[width=0.9\columnwidth]{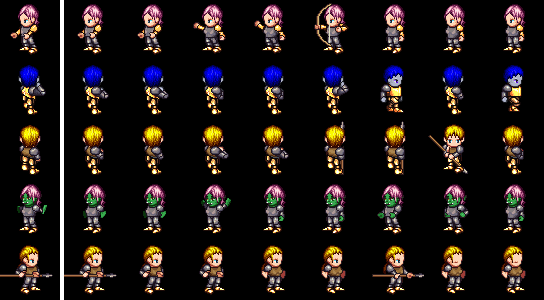}
\endminipage 
\minipage{0.40\textwidth}
\centering
\includegraphics[width=0.9\columnwidth]{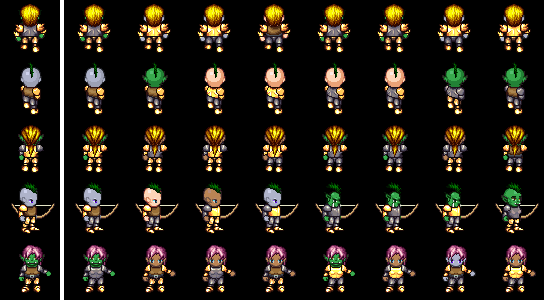}
\endminipage
\caption{left(a): sprite retrieval querying on {\it specified} component; right(b): sprite retrieval querying on {\t unspecified} component. Sprites placed at the left of the white lane are used as the query.}
\label{fig:sprite2}
\vspace{-1ex}
\end{figure}

\begin{figure}[t]
\centering
\minipage{0.40\textwidth}
\centering
\includegraphics[width=0.9\columnwidth]{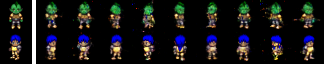}
\endminipage 
\minipage{0.40\textwidth}
\centering
\includegraphics[width=0.9\columnwidth]{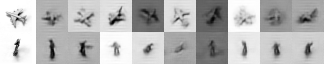}
\endminipage
\caption{left(a): sprite generation by sampling; right(b): NORB generation by sampling.}
\label{fig:sampling}
\end{figure}

\begin{figure}[h!]
\centering
\minipage{0.40\textwidth}
\centering
\includegraphics[width=0.8\columnwidth]{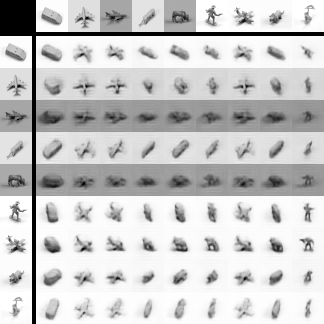}
\endminipage 
\minipage{0.40\textwidth}
\centering
\includegraphics[width=0.8\columnwidth]{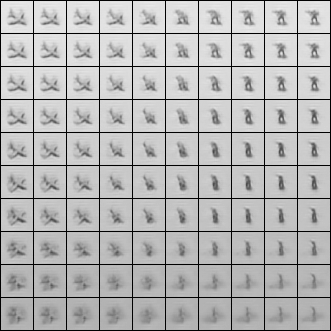}
\endminipage
\caption{left(a): A visualization grid of 2D NORB image swapping generation. Same visualization arrangement as in \ref{fig:mnist}(a); right(b): Interpolation visualization. Same arrangement as in \ref{fig:mnist}(b).}
\label{fig:norb2}
\vspace{-2ex}
\end{figure}

{\bf Sprites.} The dataset is composed of 672 unique characters (we refer to them as sprites), each of which is associated with 20 animations \cite{spritespaper}. 
%
Any image of a sprite can present 7 sources of variation: body type, gender, hair type, armor type, arm type, greaves type, and weapon type. 
%
Unlike the work in ~\cite{spritespaper}, we do not use any supervision regarding the positions of the sprites. 
The results obtained for the swapping and interpolation settings are displayed in Figure \ref{fig:sprite} while  retrieval result are showed in \ref{fig:sprite2}. Samples from the conditional model are shown in \ref{fig:sampling}(a).
We observe that the model is able to generalize to unseen sprites quite well. The generated images are sharp and single image analogies are resolved successfully, the former is an ``side-effect'' produced by the GAN term in our training loss.
The interpolation results show that one can smoothly transition between identities or positions. It is worth noting that this dataset has a fixed number of discrete positions. Thus, \ref{fig:sprite}(b) shows a reasonable coverage of the manifold with some abrupt changes. For instance, the hands are not moving up from the pixel space, but appearing gradually from the faint background.



{\bf NORB.} For the NORB dataset we used instance identity (rather than object category) for defining the labels. This results in 25 different object identities in the training set and another 25 distinct objects identities in the testing set. As in the sprite dataset, the identities used at testing have never been presented to the network at training time. 
In this case, however, the small number of identities seen at training time makes the generalization more difficult. In Figure \ref{fig:norb2} we present results for interpolation and swapping. We observe that the model is able to resolve analogies well. However, the quality of the results are degraded. In particular, classes having high variability (such as planes) are not reconstructed well. 
Also some of the models are highly symmetric, thus creating a lot of uncertainty. 
We conjecture that these problems could be eliminated in the presence of more training data.
Queries in the case of NORB are not as expressive as with the sprites, but we can still observe good behavior. We refer to these images to the supplementary material.

\begin{figure}[t]
\centering
\minipage{0.40\textwidth}
\centering
\includegraphics[width=0.9\columnwidth]{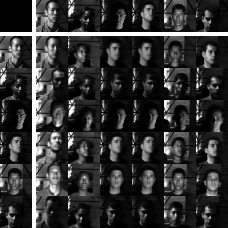}
\endminipage 
\minipage{0.40\textwidth}
\centering
\includegraphics[width=0.9\columnwidth]{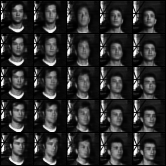}
\endminipage
\caption{left(a): A visualization grid of 2D Extended-YaleB face image swapping generation. right(b): Interpolation visualization. See \ref{fig:mnist} for description.}
\label{fig:yaleb1}
\vspace{-1ex}
\end{figure}

{\bf Extended-YaleB.} The datasets consists of facial images of 28 individuals taken under different positions and illuminations. The training and testing sets contains roughly 600 and 180 images per individual respectively. Figure \ref{fig:yaleb1} shows interpolation and swapping results for a set of testing images. Due to the small number of identities, we cannot test in this case the generalization to unseen identities. We observe that the model is able to resolve the analogies in a satisfactory, position and illumination are transferred correctly although these positions have not been seen at train time for these individuals. In the supplementary material we show samples drawn from the conditional model as well as other examples of interpolation and swapping.

{\bf Quantitative evaluation.} We analyze the disentanglement of the specified and unspecified representations, by using them as input features for a prediction task. 
We trained a two-layer neural network with 256 hidden units 
to predict structured labels for the sprite dataset, toy category for the NORB dataset (four-legged animals, human figures, airplanes, trucks, and cars) and the subject identity for Extended-YaleB dataset.
We used early-stopping on a validation set to prevent overfitting. We report both training and testing errors in Table \ref{tab:class}.
In all cases the unspecified component is agnostic to the identity information, almost matching the performance of random selection. On the other hand, the specified components are highly informative, producing almost the same results as a classifier directly trained on a discriminative manner. 
In particular, we observe some overfitting in the NORB dataset. This might also be due to the difficulty of generalizing to unseen identities using a small dataset.

\begin{table}[h]
\caption{Comparison of classification upon $z$ and $s$. Shown numbers are all error rate.}
\centering \small
\begin{tabular}{ccccccc}
\multicolumn{1}{c}{\bf set} &\multicolumn{2}{c}{\bf Sprites} &\multicolumn{2}{c}{\bf NORB} &\multicolumn{2}{c}{\bf Extended-YaleB}  \\
 & $z$ & $s$ & $z$ & $s$ & $z$ & $s$
\\ \hline \noalign{\vskip 1mm}
train & $58.6\%$ & $5.5\%$ & $79.8\%$ & $2.6\%$ & $96.4\%$ & $0.05\%$\\
test & $59.8\%$ & $5.2\%$ & $79.9\%$ & $13.5\%$ & $96.4\%$ & $0.08\%$ \\
random-chance & \multicolumn{2}{c}{$60.7\%$} & \multicolumn{2}{c}{$80.0\%$} & \multicolumn{2}{c}{$96.4\%$} \\
\end{tabular}
\label{tab:class}
\vspace{-2ex}
\end{table}

{\bf Influence of components of the framework}. It is worth evaluating the contribution of the different components of the framework. Without the adversarial regularization, the model is unable to learn disentangled representations. It can be verified empirically that the unspecified component is completely ignored, as discussed in Section~\ref{sec.cond_model}. A valid question to ask is if the training of $s$ has be done jointly in an end-to-end manner or could be pre-computed. In Section~4 of the supplementary material we run our setting by using an embedding trained before hand to classify the identities. The model is still able to learned a disentangled representations. The quality of the generated images as well as the analogies are compromised. Better pre-trained embeddings could be considered, for example, enforcing the representation of different images to be close to each other and far from those corresponding to different identities. However, joint end-to-end training has still the advantage of requiring fewer parameters, due to the parameter sharing of the encoders.

\section{Conclusions and discussion}
\label{sec.conclusions}

This paper presents a conditional generative model that learns to disentangle the factors of variations of the data specified and unspecified through a given categorization. 
The proposed model does not rely on strong supervision regarding the sources of variations. 
This is achieved by combining two very successful generative models: VAE and GAN. The model is able to resolve the analogies in a consistent way on several datasets with minimal parameter/architecture tuning. 
Although this initial results are promising there is a lot to be tested and understood.
The model is motivated on a general settings that is expected to encounter in more realistic scenarios. However, in this initial study we only tested the model on rather constrained examples. 
As was observed in the results shown using the NORB dataset, given the weaker supervision assumed in our setting, the proposed approach seems to have a high sample complexity relying on training samples covering the full range of variations for both specified and  unspecified variations. 
The proposed model does not attempt to disentangle variations within the specified and unspecified components. There are many possible ways of mapping a unit Gaussian to corresponding images, in the current setting, there is nothing preventing the obtained mapping to present highly entangled factors of variations.


{ \small
\bibliography{main}
\bibliographystyle{plain}
}

\newpage
\section*{Network architectures}

The encoder consists of a shared sub-network that splits into two separate branches. In our experiments with MNIST and the Sprites datasets, the shared sub-network is composed by three 5x5 convolutional layers with stride 2, using spatial batch normalization (BN) \cite{ioffe2015batch} and ReLU non-linearities. For the NORB and YaleB datasets, we use six 3x3 convolutional layers, with stride 2 every other layer.
The output from the top convolution layer is split into two sub-networks.
One parametrizes the approximate posterior of the unspecified component and consists of a fully-connected (FC) layer, producing two outputs corresponding to mean and variance of the approximate posterior (modeling the unspecified component). 
The other sub-network is also a fully connected used to produce the $s$ vector modeling the specified component. 
%
The decoder network takes a sample $z$ and a vector $s$ as inputs. Both codes go through a fully connected network. These representations are merged together by directly adding them and fed into a feed-forward network composed by a network mirroring encoder structure (replacing the strides by fractional strides). 
The discriminator is conditioned on the label, $id$, and configured following that used in (conditional) DCGAN.
It contains three 5x5 convolutional layers with stride 2, using BN and Leaky-ReLU with slope $0.2$. The label goes through three independent lookup tables and are added at the three first layers of representation.
The dimensionality of each representation varies from dataset to dataset. They were obtained by monitoring the results on a validation set. For MNIST, we used $16$ coefficients for each component. For sprites, NORB and Extended-YaleB, we set their dimensions as $64$ and $512$ for specified and unspecified components respectively. We found that using Stochastic Gradient Descent gives good results.

\section*{Image generation}

\begin{figure}[h!]
\centering
\minipage{0.50\textwidth}
\centering
\includegraphics[width=0.9\columnwidth]{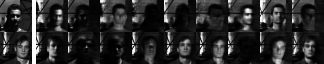}
\includegraphics[width=0.9\columnwidth]{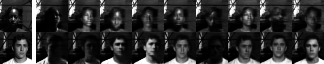}
\includegraphics[width=0.9\columnwidth]{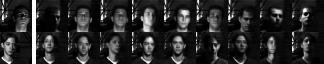}
\endminipage 
\minipage{0.50\textwidth}
\centering
\includegraphics[width=0.9\columnwidth]{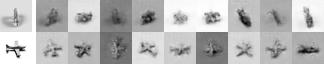}
\includegraphics[width=0.9\columnwidth]{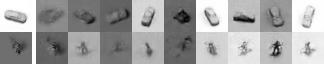}
\includegraphics[width=0.9\columnwidth]{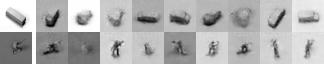}
\endminipage
\caption{More image generation. The specified part is extracted from the left images, and the unspecified part is sampled to generate the images on the right-hand side.}
\label{fig:imggenSup}
\vspace{-1ex}
\end{figure}

Figure \ref{fig:imggenSup} shows image generation. The specified part is extracted from a data sample, and an unspecified part is sampled from a Gaussian distribution. The generated sample show variation within the category of the specified part.

\section*{Interpolation}

Figure 9 shows more interpolation results. The specified and unspecified parts are extracted from two images are interpolated independently.

\begin{figure}[h!]
\centering
\minipage{0.50\textwidth}
\centering
\includegraphics[width=0.9\columnwidth]{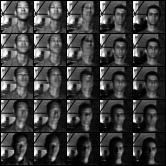}
\endminipage 
\minipage{0.50\textwidth}
\centering
\includegraphics[width=0.9\columnwidth]{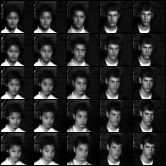}
\endminipage
\label{fig:interpSupp}
\caption{Interpolation figures, on the yaleB dataset. Only the top-left and bottom-right real faces from the test set. The the lines show interpolation along the specified part and the column show interpolation along the unspecified part. }
\vspace{-1ex}
\end{figure}

\section*{Using a pre-trained embedding}

In order to access the advantage of jointly training the system to learn the specified and unspecified parts, we tried another training scheme, summarized in the following two-step approach:
\begin{itemize}
\item Add a two-layer neural network on top of the specified part of the encoder, followed by a classification loss. Train this system in a plain supervised fashion to learn the class of the samples. When the system is converged, freeze the weights.
\item Add another encoder to produce the unspecified part of the code, and train the system as before (keeping the weights of the specified encoder frozen).
\end{itemize}
Figure 10 show the generation grid swapping the specified and unspecified parts (similar to figure 2a).

\begin{figure}[h!]
\centering
\minipage{0.40\textwidth}
\centering
\includegraphics[width=0.8\columnwidth]{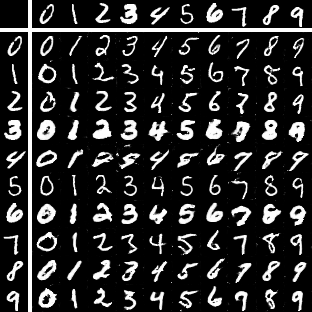}
\endminipage 
\minipage{0.40\textwidth}
\centering
\includegraphics[width=0.8\columnwidth]{mnist_swap_grid.png}
\endminipage
\label{fig:fixedS}
\caption{Swapping grid of the specified and unspecified part (see figure 2a for more details). (a) Left: pre-training the specified part of the encoder on a purely supervised task (b) Right: jointly training the whole system.}
\vspace{-1ex}
\end{figure}

\section*{Training procedure}
Algorithm \ref{algo:modeltraining} summarizes the whole training procedure. The notations are defined in sections 3 and 4 of the main paper.

\begin{algorithm}
\caption{\label{algo:modeltraining} Full model training}
\begin{algorithmic}
\FOR{number of training iterations}
\STATE{\textbf{\emph{Train the generative model}}}
\STATE{Sample a triplet of samples $(x_1,id_1)$, $(x_1', id_1)$, $(x_2, id_2)$ where $x_1$ and $x_1'$ have the same label}
\STATE{Compute the codes $(\mu_1, \sigma_1, s_1) = \mathrm{Enc}(x_1), \  (\mu_1',  \sigma_1', s_1') = \mathrm{Enc}(x_1'), \  (\mu_2, \sigma_2, s_2) = \mathrm{Enc}(x_2)$}
\STATE{Sample $z_1\sim N(\mu_1, \sigma_1), \ z_1'\sim N(\mu_1', \sigma_1'), \ z_2\sim N(\mu_2, \sigma_2)$}
\STATE{Compute the reconstructions $\tilde{X_{11}} = \mathrm{Dec}(z_1, s_1), \quad \tilde{X_{11'}} = \mathrm{Dec}(z_1, s_1')$}
\STATE{Compute the loss between $\tilde{X_{11}}$ and $X_1$, and between $\tilde{X_{11'}}$ and $X_1$, and backpropagate the gradients}
\STATE{Compute the generation $\tilde{X_{12}} = \mathrm{Dec}(z_1, s_2)$ and the adversarial loss $\log(\adv(\tilde{X_{12}}, id_2))$, and backpropagate the gradients, keeping the weights of $\adv$ frozen}
\STATE{Sample $z\sim{}N(0,1)$, generate $X_{\cdot{}2}=\mathrm{Dec}(z, s_2)$, compute the adversarial loss $\log(\adv(\tilde{X_{\cdot{}2}}, id_2))$ and backpropagate the gradients, keeping the weights of $\adv$ frozen}
\STATE{\textbf{\emph{Train the adversary}}}
\STATE{Sample a pair of samples $(x_1,id_1)$, $(x_2, id_2)$}
\STATE{Compute the codes $(\mu_1, \sigma_1, s_1) = \mathrm{Enc}(x_1),  \quad (\mu_2, \sigma_2, s_2) = \mathrm{Enc}(x_2)$}
\STATE{Sample $z_1\sim N(\mu_1, \sigma_1), \ z_2\sim N(\mu_2, \sigma_2)$}
\STATE{Compute the reconstructions $\tilde{X_{11}} = \mathrm{Dec}(z_1, s_1), \quad \tilde{X_{12}} = \mathrm{Dec}(z_1, s_2)$}
\STATE{Compute the adversarial loss (negative sample) $\log(1-\adv(X_{12}, id_2))$ and backpropagate the gradients, keeping the weights of Enc and Dec frozen}
\STATE{Compute the adversarial loss (positive sample) $\log(\adv(X_2, id_2))$ and backpropagate the gradients, keeping the weights of Enc and Dec frozen}
\ENDFOR
\end{algorithmic} 
\end{algorithm}

\end{document}